\documentclass[acmsmall]{acmart}

\AtBeginDocument{%
  }
\usepackage{siunitx}
\usepackage{natbib}
\usepackage{multirow}
\usepackage{wrapfig}
\DeclareSIUnit{\nothing}{\relax}
\setcopyright{acmlicensed}
\copyrightyear{2025}
\acmYear{2025}
\acmDOI{10.1145/3719210}

\acmJournal{JACM}
\acmVolume{1}
\acmNumber{1}
\acmArticle{1}
\acmMonth{1}

\usepackage{mathtools}
\DeclarePairedDelimiter\floor{\lfloor}{\rfloor}
\DeclareMathOperator{\arctantwo}{arctan2}

\definecolor{somegray}{rgb}{0.5, 0.5, 0.5}
\newcommand{\darkgrayed}[1]{\textcolor{somegray}{#1}}
\makeatletter
\newcommand*\titleheader[1]{\gdef\@titleheader{#1}}
\AtBeginDocument{%
  \let\st@red@title\@title
  \def\@title{%
    \vskip-2.0em
    \bgroup\normalfont\large\centering\@titleheader\par\egroup
    \vskip0.0em\st@red@title}
}

\makeatother

\titleheader{\darkgrayed{This paper has been accepted for publication in the 2025 ACM Journal on Autonomous Transportation Systems\\\copyright 2025 ACM.}}

\usepackage[firstpage]{draftwatermark} 
\SetWatermarkText{This paper has been accepted for publication in the 2025 ACM Journal on Autonomous Transportation Systems\\\copyright 2025 ACM.} 
\SetWatermarkScale{0.075} 
\SetWatermarkColor[gray]{0.85} 
\SetWatermarkAngle{0} 
\SetWatermarkVerCenter{0.1\paperheight} 

\begin{document}

\title[An Efficient Ground-aerial Transportation System for Pest Control Enabled by Autonomous AI-based Nano-UAVs]{An Efficient Ground-aerial Transportation System for Pest Control Enabled by AI-based Autonomous Nano-UAVs}





\author{Luca Crupi}
\email{luca.crupi@supsi.ch}
\affiliation{%
  \institution{IDSIA, SUPSI}
  \city{Lugano}
  \country{Switzerland}
}
\author{Luca Butera}
\email{luca.butera@usi.ch}
\affiliation{%
  \institution{IDSIA, USI}
  \city{Lugano}
  \country{Switzerland}
}
\author{Alberto Ferrante}
\email{alberto.ferrante@usi.ch}
\affiliation{%
  \institution{IDSIA, USI}
  \city{Lugano}
  \country{Switzerland}
}
\author{Alessandro Giusti}
\email{alessandro.giusti@usi.ch}
\affiliation{%
  \institution{IDSIA, SUPSI}
  \city{Lugano}
  \country{Switzerland}
}
\author{Daniele Palossi}
\email{daniele.palossi@supsi.ch}
\affiliation{%
  \institution{IDSIA, SUPSI}
  \city{Lugano}
  \country{Switzerland}
}
\affiliation{%
  \institution{IIS, ETH Z\"urich}
  \city{Z\"urich}
  \country{Switzerland}
}



\begin{abstract}

Efficient crop production requires early detection of pest outbreaks and timely treatments; we consider a solution based on a fleet of multiple autonomous miniaturized unmanned aerial vehicles (nano-UAVs) to visually detect pests and a single slower heavy vehicle that visits the detected outbreaks to deliver treatments.
To cope with the extreme limitations aboard nano-UAVs, e.g., low-resolution sensors and sub-\SI{100}{\milli\watt} computational power budget, we design, fine-tune, and optimize a tiny image-based convolutional neural network (CNN) for pest detection. 
Despite the small size of our CNN (i.e., \SI{0.58}{\giga Ops/inference}), on our dataset, it scores a mean average precision (mAP) of 0.79 in detecting harmful bugs, i.e., 14\% lower mAP but 32$\times$ fewer operations than the best-performing CNN in the literature. 
Our CNN runs in real-time at~\SI{6.8}{frame/\second}, requiring~\SI{33}{\milli\watt} on a GWT GAP9 System-on-Chip aboard a Crazyflie nano-UAV. 
Then, to cope with in-field unexpected obstacles, we leverage a global+local path planner based on the A* algorithm. 
The global path planner determines the best route for the nano-UAV to sweep the entire area, while the local one runs up to~\SI{50}{\hertz} aboard our nano-UAV and prevents collision by adjusting the short-distance path. 
Finally, we demonstrate with in-simulator experiments that once a 25 nano-UAVs fleet has combed a 200$\times$\SI{200}{\meter} vineyard, collected information can be used to plan the best path for the tractor, visiting all and only required hotspots. 
In this scenario, our efficient transportation system, compared to a traditional single-ground vehicle performing both inspection and treatment, can save up to~\SI{20}{\hour} working time.

\end{abstract}





\maketitle

\begin{acks}
We thank Hanna Müller, Victor Kartsch, and Luca Benini, authors of~\cite{müller2024gap9shield150gopsaicapableultralow} for providing us with a GAP9Shield prototype.
\end{acks}

\section{Introduction} \label{sec:intro}

Heterogeneous ground-aerial autonomous robots combine the advantages of both (i.e., agile flying drones and high-payload ground vehicles) and mitigate the limits of each, i.e., reduced payloads and exploration speed respectively, leading to efficient autonomous transportation systems~\cite{chatziparaschis2020aerial, 8955969}.
Many different transportation applications can benefit from this pattern, from warehouse management~\cite{inproceedings_warehouse} to traffic monitoring and control~\cite{rodic2012ambientally}, precision agriculture~\cite{9177181}, and emergency response scenarios~\cite{rajapakshe2023coll}.
In all these cases, combining accurate analysis (e.g., from a flying drone) with decentralized decision-making processes (e.g., optimal path planning) leads not only to the obvious economic advantage of reducing costs (e.g., transportation and delivery costs) but also to additional social and environmental benefits, of saved time and resources.

This work addresses smart farming crop production, enabling prompt, precise, and efficient treatments in cultivated fields.
To reach the goal of timely fine-grained treatments (up to single plants) in the event of pest outbreaks, accurate pest detection and optimal planning of routes for slow machinery are of the essence.
Reducing the use of pesticides not only brings economic benefits for farmers but also reduces the environmental impact of mass production.
This work addresses this challenge by designing a two-level autonomous transportation system.
The forefront is represented by a fleet of miniaturized autonomous unmanned aerial vehicles (UAVs), a new class of inexpensive airborne called nano-UAVs~\cite{9474262, lamberti2024sim} as big as the palm of one hand, i.e., sub-\SI{50}{\gram} weight and less than \SI{10}{\centi\meter} in diameter.
Instead, on the ground, we rely on an automated tractor acting as the backbone for the heavy-duty job~\cite{applmech3030049, tahmasebi2022autonomous}, i.e., pest control in the environment.
Thanks to their agility, nano-UAVs can quickly act as probes to comb vast cultivated areas, looking for the first signs of pest outbreaking, leaving the fine-grained treatment (e.g., spraying chemicals) to the bulky and slow tractor.

However, the agility of nano-UAVs comes with ultra-constrained onboard resources, such as simple sensors, a few tens of \SI{}{\mega\byte} of memory, and sub-\SI{100}{\milli\watt} for the onboard computational power~\cite{palossi2019open}.
Nevertheless, our nano-UAVs are required to \textit{i}) analyze, in real-time, images acquired on board for detecting harmful insects, and \textit{ii}) fly autonomously, following a pre-defined path but avoiding collisions with unexpected dynamic obstacles.
To cope with this scenario, we employ a commercial off-the-shelf (COTS) Crazyflie nano-UAV featuring an STM32 microcontroller unit (MCU) and extended with a low-resolution camera, an 8$\times$8 Time-of-Flight (ToF) depth sensor, and an ultra-low power Greenwaves Technologies (GWT) GAP9 multi-core System-on-Chip (SoC).

We address the former task by optimizing and deploying a State-of-the-Art (SoA) vision-based deep learning model, i.e., SSDLite-MobileNetV3~\citep{howard2019searching}, re-trained for pest detection and running on the GAP9 SoC.
Instead, the latter task is handled with a \textit{global+local path planning} approach~\cite{marin2018global}, where both planners are based on the A* algorithm~\cite{hart1968formal}, a graph-based path planning algorithm.
The vast cultivated field is split into 40$\times$\SI{40}{\meter} areas due to the limited battery lifetime of nano-UAVs (i.e., $\sim$\SI{6}{\minute}).
For each area, we convert a static 2D occupancy map~\cite{30720} of the field (i.e., a vineyard in our case) in an 8-connected graph~\cite{HALIN1969150}.
Before the mission starts, the global path planner computes the optimal path for each nano-UAV on such a graph, enforcing the complete exploration of the area.
Then, during the mission, the local planner, running on the STM32 aboard our nano-UAVs, takes in input a local occupancy map of 4$\times$\SI{4}{\meter} continuously updated from the onboard depth sensor.
Therefore, the local planner can adjust the path if a new obstacle lies on the pre-computed global path.
Thanks to the real-time performance of the local planner up to \SI{50}{\hertz}, we achieve the desired reactive obstacle avoidance functionality.

Our main contribution is a novel integrated transportation system that combines autonomous ground and aerial vehicles. 
We design, optimize, and deploy a SoA convolutional neural network (CNN) capable of achieving up to 0.79 mean average precision (mAP) for accurate pest detection, along with a local path planner for obstacle avoidance, both implemented on the two MCUs of our nano-UAV. 
We conduct a thorough evaluation of the CNN’s classification and inference performance, including power consumption measurements on the GAP9 SoC, which operates within \SI{33}{\milli\watt} while achieving a throughput of \SI{6.8}{frame/\second}. 
Additionally, we implement and validate the complete transportation system in the Webots\footnote{https://cyberbotics.com/} simulator, including sensor interfacing and executing the global and local path planning algorithms.
Our findings demonstrate that the proposed ground-aerial transportation system significantly outperforms traditional methods relying only on a single tractor for pest detection and treatment by reducing exploration time up to \SI{20}{\hour}, assuming an area of 200$\times$\SI{200}{\meter} and a mean speed of \SI{0.2}{\meter/\second}~\cite{9177181}.
\section{Related works}

In this section, we first go over insect detection and control and then focus on the problem of routing in autonomous vehicles.

\subsection{Pest detection and control}


\begin{table*}[t]
    \small
    \caption{Network survey (number of parameters, operations, throughput, and mAP) on the models used in~\cite{9601235} and extended with an additional one introduced in our work (\textbf{in bold}). The GPU employed is an Nvidia GeForce RTX 2080. The number of operations is measured with multiply-accumulate operations (MACs).}
    \label{tab:other_models_params}
    \resizebox{\linewidth}{!}{%
    \begin{tabular}{llcccccc}
    \toprule
    \textbf{Network} & \textbf{Input size} & \textbf{\# Param. [\SI{}{\mega\nothing}]} & \textbf{\# Op. [\SI{}{\giga MAC}]} & \textbf{Device} & \textbf{Frame-rate [\SI{}{\hertz}]} & \textbf{mAP}\\
    \toprule
    FasterRCNN-VGG16-FPN & $800\times800\times3$ & 31.90 & 275.23 & GPU & 11.90 & 0.92\\
    FasterRCNN-ResNet101-FPN & $800\times800\times3$ & 60.20 & 167.41 & GPU & 11.23 & 0.92\\
    FasterRCNN-DenseNet169-FPN & $800\times800\times3$ & 30.00 & 73.99 & GPU & 7.59 & 0.91\\
    FasterRCNN-MobileNetV3-FPN & $800\times800\times3$ & 18.90 & 18.41 & GPU & 60.92 & 0.93\\
    \midrule
    RetinaNet-VGG16-FPN & $800\times800\times3$ & 22.90 & 270.06 & GPU & 12.37 & 0.91\\
    RetinaNet-ResNet101-FPN&$800\times800\times3$&51.20&174.85 &GPU&11.83&0.93\\
    RetinaNet-DenseNet169-FPN&$800\times800\times3$&21.00&65.10 &GPU&18.49&0.92\\
    RetinaNet-MobileNetV3-FPN&$800\times800\times3$&10.60&11.09 &GPU&48.91&0.91\\
    \midrule
    SSD-VGG16&$300\times300\times3$&12.10&26.58 &GPU&42.58&0.56\\
    SSD-ResNet101&$300\times300\times3$&32.30&42.39 &GPU&29.76&0.87\\
    SSD-DenseNet169&$300\times300\times3$&11.80&24.51&GPU&28.59&0.92\\
    SSD-MobileNetV3&$300\times300\times3$&7.50&3.42&GPU&33.20&0.80\\
    \midrule
    \textbf{SSDLite-MobileNetV3}&\textbf{$\mathbf{320\times240\times3}$}&\textbf{3.44}&\textbf{0.58}&\textbf{GAP9}&\textbf{6.8}&\textbf{0.79}\\
    \bottomrule
    \end{tabular}
    }
\end{table*}

Traditional techniques to avoid spreading dangerous insects in cultivated fields have employed coarse-grained use of pesticides spraying entire fields, with the downside of potential health issues, environmental impact~\cite{DLEntomology}, and waste of resources.
More fine-grained control of pest outbreaks can be achieved by relying on humans to visually inspect crops and manually treat only where needed, but this approach is expensive and potentially dangerous for human operators.
An important step forward in the precise treatment of pests has been achieved by introducing insect traps, originally monitored by humans~\cite{preti2021insect} and later on, thanks to the advent of battery-powered embedded systems featuring automatic bug detection capabilities.
This novel technology, also called \textit{smart traps}~\cite{SmartTrap}, has been successfully utilized against the Cydia pomonella~\cite{JU2021104925} and the Popillia japonica~\cite{EPPO}, leaving the final chemical treatment either to humans~\cite{preti2021insect} or by autonomous ground robots~\cite{tahmasebi2022autonomous}.

SoA insect detectors, employed in combination with smart traps, leverage image-based computationally intensive deep learning models~\cite{FasterRCNNInsect2, RetinaNetInsect1, RetinaNetInsect1, RFCNInsect1}, as surveyed by~\cite{PestDetectionSurvey}.
Table~\ref{tab:other_models_params} presents a broad overview of SoA models for pest detection, based on~\citet{9601235}, with the addition (last row) of our proposed model, which is based on a MobileNet CNN~\citep{sandler2018mobilenetv2, howard2019searching}.
As surveyed in Table~\ref{tab:other_models_params}, SoA algorithms require a significant amount of computational and memory resources in the order of high-end GPUs.
For this reason, smart traps rely on remote powerful servers to perform pest detection~\cite{agriculture12101745} on streamed images, which \textit{i}) hinges on the power consumption of smart trap due to radio transmission, \textit{ii}) requires additional infrastructure, e.g., 4G/5G radio modules, and \textit{iii}) raises cyber-security concerns on the communication channel, requiring further complexity to implement robust security features. 
These limitations can be overcome using standalone smart traps or autonomous nano-UAVs to perform pest detection, but introduce the challenge of accurate detection within the limited power envelope of MCU-class processors, i.e., sub-\SI{100}{\milli\watt} computational power.
Finally, in the case of autonomous nano-UAVs
However, nano-UAVs have the additional requirements of real-time onboard inference, as we want to explore the largest area possible in their limited lifetime, i.e., $\sim$\SI{6}{\minute}, meanwhile inspecting the cultivated field.

Recent works allow the deployment of object detection algorithms on the edge.
Squeezed edge YOLO~\citep{humes2023squeezed} performs object detection within less than 1 million parameters and can be deployed on parallel ultra-low power SoC.
The work by~\citet{rusci2023parallel} introduces a Viola-Jones-based algorithm for detecting Cydia pomonella.
This algorithm runs on a battery-powered embedded system mounted on sticky pad-based traps.
Despite the simplicity of this approach, which uses pre-trained visual patches on a reduced amount of data, it peaks at \SI{2.5}{frame/\second} on a GWT GAP8 SoC. 
The approach, however, does not allow for the classification of different insects, which is part of our task.
\citet{BETTISORBELLI2023108228} adopts a YOLO network to detect insects in aerial UAV images. 
The method is accurate (up to 0.92 of mAP), but struggles with clusters of insects, as non-maximum suppression removes detections with high intersection over union.
Finally, \citet{10137154} presents an SSD-MobileNetV2 object detection algorithm fully deployed on the GAP8 SoC running aboard an autonomous nano-UAV to detect tin cans and bottles.
This system reaches up to \SI{1.6}{frame/\second} with an mAP of 0.5.
Similarly, in the work by~\citet{9401730}, a GAP8 SoC is employed for automatic license plate recognition, using a MoileNetV2 with an SSDLite detector.
In this work, we employ MobileNets~\cite{howard2017mobilenets} and pair them with an SSDLite detector head on the GAP9 SoC, achieving a peak throughput of~\SI{6.8}{frames/second}. 
This approach enables precise insect detection and classification on resource-constrained devices, making it suitable for deployment on autonomous nano-UAVs.
As such, w.r.t. currently available object detection algorithm running on resource-constrained platforms that can fit the power envelope of our nano-UAV, we improve the mAP by $\sim$0.3, reaching an mAP of 0.79 with our onboard running CNN, allowing real-world applicability of our system.

\subsection{Routing for autonomous transportation systems}

In our scenario, vehicle routing must be addressed at two different levels: global and local.  
Global routing consists of defining the overall path to follow in order to completely explore an area or visit a set of points of interest, while minimizing some cost metric. 
This problem is widely known in literature as the traveling salesman problem (TSP).  
Local routing, instead, consists of reactive planning of a local path, following some goal while avoiding unexpected obstacles the vehicle might encounter.  
The work by~\citet{marin2018global}, similarly to ours, presents a partitioning of the routing problem between global routing, for the overall path, and local routing, for obstacle avoidance. 
However, they focus on a setting with a single land vehicle.

\paragraph{Global routing}
The TSP is a well-known optimization problem: the objective is to find the route the salesman has to travel to visit a set of destinations, represented as nodes in a graph, which minimizes some cost metrics. 
Our global routing scenario falls within the category of multiple traveling salesman problem (MTSP), which is a generalization of TSP to the multiple agent setting~\cite {cheikhrouhou2021comprehensive}, generally cooperative.
In this scenario, common metrics to minimize are the total path length among agents (Min-Sum) or the maximum path length (Min-Max). 
Variations of MTSP may involve multiple depots from which the salesman can depart or arrive, as well as constraints on departing and arriving at the same depot. 
Moreover, specific applications may adopt additional optimization objectives, like travel time or resource consumption minimization. 

In some cases (e.g., UAVs), additional constraints might be considered, like fuel or physical maneuverability. 
The MTSP problem can be decomposed into a node assignment part, in which each node in the exploration graph is assigned to a specific agent, and a path planning part, in which the path over the assigned sub-graph is computed independently for each agent. 
Exact methods for the multiple depots MTSP have been proposed~\citep{oberlin2009transformation, sundar2017algorithms}. 
However, these are computationally expensive in general. 
The most widely adopted approaches are so-called meta-heuristics, which use optimization algorithms to find near-optimal solutions with less computational demands. 
In this category, evolutionary approaches, such as genetic algorithms (GA), have been widely adopted~\citep{al2019comparative, bolanos2015multiobjective, yuan2013new}, while also particle swarm optimization (PSO)~\citep{wei2020particle}, ant colony optimization (ACO)~\citep{lu2019mission} and artificial bee colonies (ABC)~\citep{venkatesh2015two} have been proposed to address the task. 

Market-based approaches have also been explored to a lesser extent~\citep{kivelevitch2013market, elango2011balancing, koubaa2017move}. 
These techniques treat the optimization process as an auction, in which tasks (nodes) are assigned based on the agent's bidding. 
In recent years, research on the MTSP has moved to take UAVs into account.
\citet{wichmann2015smooth} proposed an approach based on location clustering for task assignment and GA to path definition in the context of wireless sensory networks and mobile sinks. 
\citet{hayat2017multi}, proposes a classical GA algorithm with a time minimization objective for UAVs in search and rescue missions, while~\citet{du2017precision} proposes GA combined with a hierarchical approach to solving the planning task in the context of pesticide spraying in precision agriculture.
\citet{chen2019multi} specifically addresses the problem of energy constraints in UAVs, in the context of coverage path planning, using a modified GA which explicitly considers the resource limitations of such vehicles. 
Conversely, \citet{ma2019coordinated} proposed a routing algorithm that accounts for explicit constraints in task execution time (i.e., travel time) by using task clustering and a modified GA for constrained path planning. 
\citet{hu2020reinforcement}, differently from previous work, proposes the usage of reinforcement learning (RL) for task (node) assignment, while single path planning can be handled by any available solver.  

\paragraph{Local routing}
Local routing consists of solving the problem of finding an optimal local path (i.e., a path relative to the robot's vicinity w.r.t. its speed and distance from objects), which minimizes deviation from the global path, while avoiding collisions. 
Usually, this is implemented by minimizing some cost function that accounts for real-time information coming from onboard sensors (e.g., LIDAR, sonar, and cameras).  
Overall, these algorithms modify the robot's planned path to implement reactive behavior in the autonomous vehicle while still reaching some goal.  
\citet{marin2018global}, adopts the time elastic band (TEB) algorithm~\citep{rosmann2013efficient}, which formulates the local routing problem as finding the optimal sub-path in a dynamic hyper-graph which jointly represents the global path (as a sequence of vehicle states) and feasibility constraints (e.g., velocity, obstacles). 
The optimization problem is then solved by a non-linear programming solver, and the hyper-graph's sparsity plays an important role in terms of computation time.  

\citet{hossain2022local} propose a local routing algorithm based on combining the dynamic window approach (DWA)~\citep{fox1997dynamic}, for optimal velocity planning, with an improved follow the gap method (FGM)~\citep{sezer2012novel}, for obstacle avoidance.  
DWA tries to find the optimal velocity, given the robot's physical constraints, to reach the goal while avoiding static obstacles, resulting in non-reactive behavior.  
FGM, on the other hand, determines the optimal approach angle to avoid collision by fitting in the largest gap between observed obstacles. 
This approach is strongly geared towards safety rather than optimal task execution time, i.e., it favors avoiding collisions, even if it means selecting a longer path.
\citet{khatib1985real} originally proposed the artificial potential fields (APF) approach, which models goals as attractive fields and obstacles as repulsive fields guiding the robot. 
This approach, however, has difficulties handling local minima and is not guaranteed to reach the goal.  
Another known approach to local routing is that of vector field histograms (VFH)~\citep{borenstein1991vector, ulrich1998vfh+, ulrich2000vfh}, which statistically models the surrounding environment in the form of histogram grids, whose values are based on sensor readings. 
This approach then selects the local path to go towards regions of low collision probability that are headed in the goal's direction.  

In choosing routing algorithms, we must consider two peculiar aspects of our setting: UAVs will always explore the same area, while land robots will explore different areas based on the information provided by UAVs. 
Given this, exact algorithms can be used for global routing, as computation is sporadic and can be carried out on unconstrained devices.
Conversely, to avoid unexpected obstacles, such as workers or animals, which are not known when the global path is computed, we need a local planner running in real time onboard the robots.
This demands short execution time, thus heuristic techniques that do not explore all possible paths.

In this work, we are focusing on proposing a novel combination of known concepts (e.g., object detection and autonomous transportation) to enable optimized, fully autonomous pest control in the open field.
We adopt the classic A*~\citep{hart1968formal} algorithm for both global and local routing.
A* is a graph-based path planning algorithm that has a computational cost dependent on the cost metrics used to compute the path.
The cost metric that we use, i.e., Euclidean distance and the Euclidean distance weighted with the type of destination node (as explained in Section~\ref{subsec:routing}), are lightweight and, as such, allow the real-time computation onboard each \SI{50}{\gram} nano-UAV of our swarm in the local routing version, thus, using the 4$\times$\SI{4}{\meter} local occupancy map.
Our approach provides a solution for obstacle avoidance that can run on resource-constrained nano-UAVs in real-time, allowing the detection of static and dynamic obstacles with maximum speeds that range from~\SI{4.5}{\meter/\second} to ~\SI{19.5}{\meter/\second} depending on the dimension and distance of the obstacle.
To the best of our knowledge, this work provides the first graph-based solution for obstacle avoidance on nano-UAVs relying only on onboard computations.

Furthermore, our work also provides novel solutions if we consider the complete transportation system composed of the nano-UAVs and the ground robot. 
The only prior work that considers the task of carrying out pest control using autonomous vehicles is~\citet{du2017precision}. 
In such work, standard-sized quadcopters follow a pre-computed route, optimized offline, to spray an area with pesticides. 
Even though our specific task differs, our approach's innovations can be transferred to the setting in~\citet{du2017precision}. 
In particular, instead of spraying an entire area with pesticides, we autonomously detect locations that require intervention using fast and effective nano-UAVs. 
The actual intervention is carried out only when and where it is required by a large and slow ground vehicle, resulting in a drastic reduction of time, particularly for early detected insect hotspots.

\section{System implementation}

In this section, we first introduce the nano-UAV platform used in our work, along with the Webots simulation employed.
Afterward, we discuss the deep learning models we port to our platform for the detection of pest insects and the routing algorithms we adopt for path planning.

\subsection{Robotic platform} \label{subsec:robotic_platforms}

\begin{figure*}[tb]
\centering
\includegraphics[width=1.0\linewidth]{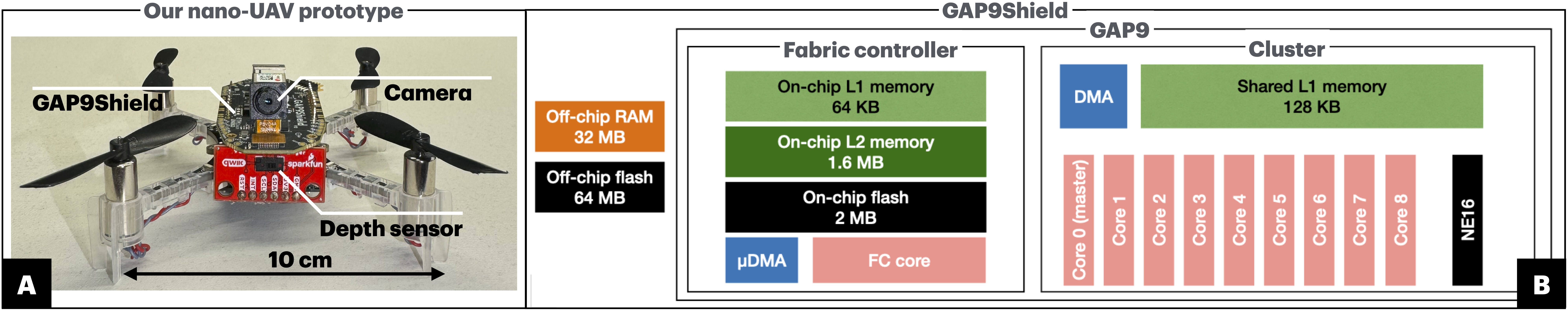}
\caption{Picture of our nano-UAV prototype (A) and the GAP9Shield block diagram (B).}
\label{fig:gap9drone}
\end{figure*}

\begin{wrapfigure}{R}{0.5\textwidth}
  \begin{center}
    \includegraphics[width=0.48\textwidth]{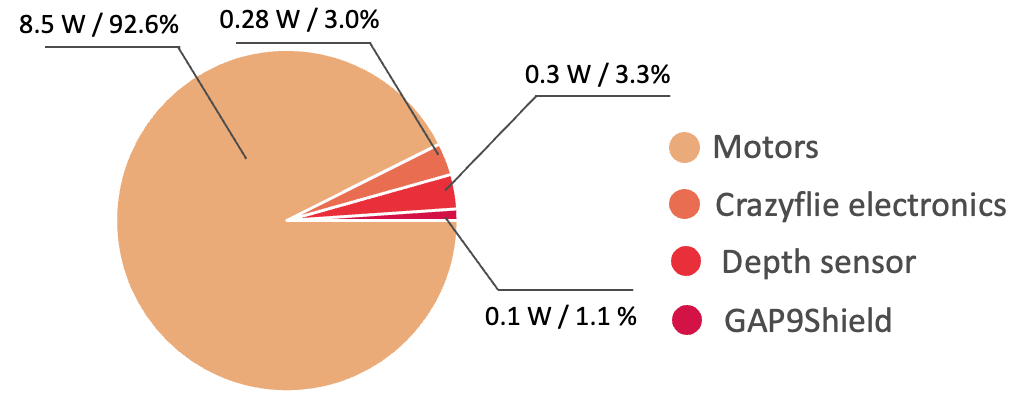}
  \end{center}
  \caption{\label{fig:pie_chart_power}Power breakdown of our robotic platform.}
\end{wrapfigure}

\begin{wrapfigure}{tb!}{0.5\textwidth}
  \begin{center}
    \includegraphics[width=0.48\textwidth]{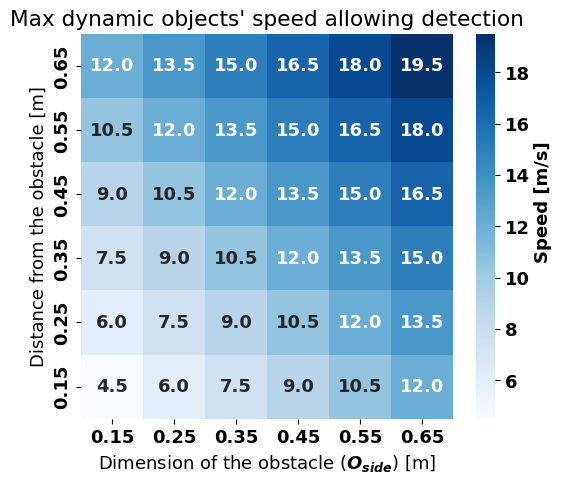}
  \end{center}
  \caption{\label{fig:heatmap_distances_speed}The maximum detectable object speed depends on both the object dimension ($O_\text{side}$) and the distance from the sensor. $O_\text{side}$ is defined as the projection of the object on the sensor frame.}
\end{wrapfigure}

Our work relies on a Crazyflie 2.1\footnote{https://www.bitcraze.io/products/crazyflie-2-1-plus} nano-UAV for the exploration of the vineyard that consumes, on average, evaluated with wind speeds between 0.5 and \SI{3.4}{\meter/\second}, $\sim$\SI{8.8}{\watt}~\cite{9811834}, including the electronics.
The UAV features an STM32 MCU, providing auto-pilot functionalities and acting as an interface for peripherals and sensors, that requires up to \SI{138}{\milli\watt}@\SI{168}{\mega\hertz}.
Basic functionalities involve state estimation and a low-level proportional-integral-derivative (PID) controller.
The UAV has two expansion boards, as detailed in Figure~\ref{fig:gap9drone}-A, in addition to the flow deck: a custom-made board that connects the STM VL53LC5CX\footnote{\href{https://www.st.com/resource/en/datasheet/vl53l5cx.pdf}{https://www.st.com/resource/en/datasheet/vl53l5cx.pdf}} ToF depth sensor to the Crazyflie 2.1 via I2C bus; and a GAP9Shield~\cite{müller2024gap9shield150gopsaicapableultralow}.
We report in Figure~\ref{fig:pie_chart_power} the power breakdown of our robotic platform.
The ToF depth sensor provides 8$\times$8 depth maps at~\SI{15}{\hertz} while consuming~\SI{313}{\milli\watt}. 
Each pixel of the 8$\times$8 depth map represents an independent reading of depth in a square field-of-view (FoV) with a \SI{63}{\degree} diagonal; depth ranges between 0 and~\SI{4}{\meter}.
The sensor specification reports an accuracy of $\pm15$\si{\milli\meter} for readings in the~\SI{20}{\milli\meter} to~\SI{20}{\centi\meter} range, while from~\SI{20}{\centi\meter} to~\SI{4}{\meter} the error grows to $\pm11\%$. 
The framerate of the depth sensor provides an upper limit to the objects' speed to guarantee its detectability.  
We report in Figure~\ref{fig:heatmap_distances_speed} the maximum detectable speed, which ranges between \SI{4.5}{\meter/\second} and \SI{19.5}{\meter/\second}, that depends on both the distance from the sensor and the object side projected on the depth sensor frame.
We consider distances up to~\SI{0.65}{\meter} since it is the typical working distance of the depth sensor in the case of gray targets, with 17\% reflectance, which represents a worst-case scenario for our depth sensor.

The GAP9Shield provides onboard intelligence enabled by a RISC-V-based GAP9\footnote{https://greenwaves-technologies.com/gap9\_processor} SoC together with an RGB OmniVision OV5647 camera with up to \SI{45}{frame/\second} @ VGA image resolution.
The SoC, depicted in Figure~\ref{fig:gap9drone}-B, features ten cores equipped with a single-precision floating point unit (FPU).
The cores are logically split into two domains, i.e., the \textit{fabric controller} (FC) that features one core and the \textit{cluster} (CL) with nine cores. 
The FC orchestrates the work for the multi-core cluster and provides an interface with external peripherals, while the CL is particularly suitable for CNNs since it can handle general-purpose intense parallel workloads.
The GAP9Shield has a four levels memory hierarchy which is composed of two on-chip memories, an L1 memory of~\SI{128}{\kilo\byte}, and an L2 memory of~\SI{1.6}{\mega\byte} while the~\SI{32}{\mega\byte} RAM and the~\SI{64}{\mega\byte} FLASH are hosted off-chip.
The GAP9 SoC includes an \textit{int8} hardware accelerator, the NE16, specifically tailored for linear algebra computations that can, among others, accelerate convolutions execution.

\subsection{Insect detector}

To address insect detection and classification, we utilize a CNN that we designed~\cite{crupi2024deep} to fit the computational constraints of the GAP9. 
Specifically, we focus on fine-tuning, testing, and deploying a MobileNet-based architecture~\cite{howard2019searching} that was originally pre-trained on the COCO dataset~\cite{cocodataset}. 
Fine-tuning, a well-known method in transfer learning~\cite{TransferLearning}, involves adapting a model trained on one task -- such as image classification on the COCO dataset~\cite{cocodataset} -- to a similar task in a different domain while taking advantage of the general features learned in the original training.

\begin{wrapfigure}{R}{0.5\textwidth}
  \begin{center}
    \includegraphics[width=0.48\textwidth]{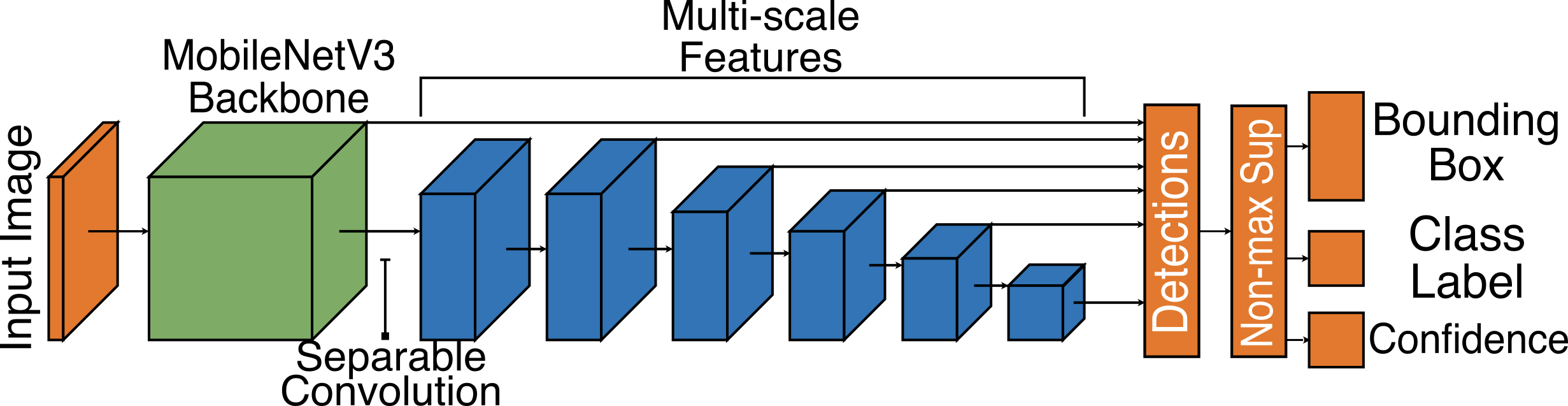}
  \end{center}
  \caption{\label{fig:ssdlite_mnet}SSDLite with MobileNetV3 backbone architecture.}
\end{wrapfigure}

As Figure~\ref{fig:ssdlite_mnet} reports, we selected the MobileNetV3 architecture combined with an SSDLite detector, which is a variant of the MobileNetV3 with SSD, as proposed in~\cite{9601235} for edge device deployment. 
The backbone of the network is a MobileNetV3~\cite{howard2019searching} model, consisting of~\SI{3.44}{\mega\nothing} parameters and requiring~\SI{584}{\mega\nothing MAC} per inference, similar to the architecture described in~\cite{9601235}. 
The SSDLite detector, serving as the network’s head, modifies the standard SSD by replacing regular convolutions with depth-wise separable convolutions. 
This change significantly reduces computational cost -- up to 3.3$\times$ -- without affecting the mAP, as shown in Table~\ref{tab:other_models_params}. 
The SSDLite detector also includes a non-maximal suppression layer to filter out redundant or low-confidence predictions. 
This model takes a 320$\times$240 pixel image as input and outputs multiple bounding boxes along with class labels and confidence scores. 
We fine-tune the network over 300 epochs using an SGD optimization algorithm with a learning rate of 0.00025. 
The training process follows a learning rate schedule with a momentum of 0.9, a weight decay of 0.0005, a warmup phase of 10 epochs, and a minimum learning rate of 0.00005.

\subsubsection{Dataset}

\begin{wrapfigure}{t}{0.5\textwidth}
  \begin{center}
    \includegraphics[width=0.48\textwidth]{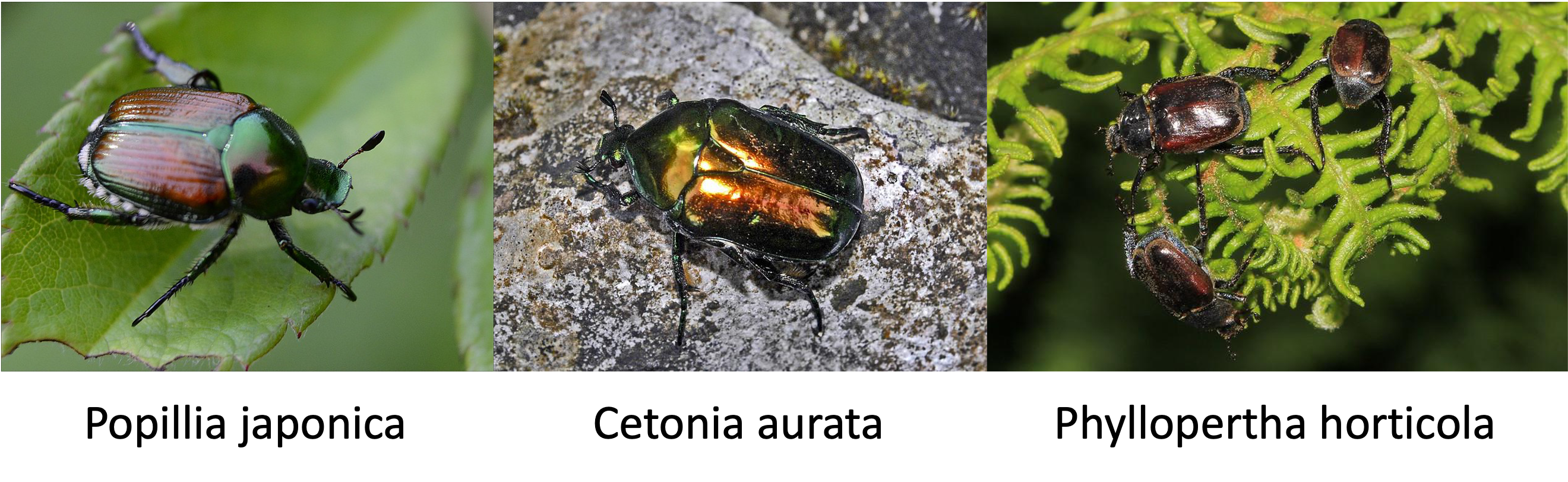}
  \end{center}
  \caption{\label{fig:datasetSamples}Samples of the three classes in our dataset~\citep{9601235}.}
\end{wrapfigure}

Since our task is insect detection, we adopt the dataset from~\citet{9601235}, which includes more than 3,300 images of insects, spanning three classes: \textit{Popillia japonica}, \textit{Cetonia aurata}, and \textit{Phyllopertha horticola}. 
The first is a dangerous pest insect, while the other two, often mistaken for the first, are not.
In total, the dataset contains 1,422, 1,318, and 877 samples for each class, respectively.
Note that the dataset is the result of a strict filtering process starting from a collection of more than 36,000 images gathered through the Internet. 
This ensures the near-absence of duplicates that could harm model evaluation.
Even though the dataset is not publicly available, a similar dataset can be built by means of popular image search engines, e.g., Google, Flickr, using the 3 insect names as search keys.
Figure~\ref{fig:datasetSamples} shows examples of the three insect classes.
The dataset was split into training and testing data, with an 80-20\% ratio.

\subsubsection{Deployment}

To deploy the CNN, we use a platform-specific tool, NN-Tool, developed by GWT, that converts Python code into C code tailored for the GAP9 SoC.
NN-Tool supports deployment from \textit{onnx} or \textit{tflite} files. 
In our case, we fine-tune the MobileNetV3 with the SSDLite detector (pre-trained on COCO in PyTorch) on our dataset to produce a \textit{float32} CNN, which is then converted into an \textit{onnx} file. 
To ensure compatibility with NN-Tool, non-maximal suppression is implemented outside of the neural model, as it is not supported by the tool. 
NN-Tool also handles tensor tiling and memory hierarchy management, optimizing usage across all on-chip memory levels. 
As the GAP9 has a single-precision FPU, we deploy the network in float16 and int8 formats.
Additionally, we utilize the NE16 hardware accelerator for CNN applications, which requires CNNs quantized in int8. 
Thus, we deploy on the GAP9 SoC three versions of our CNN: float16, int8, and int8 with NE16 acceleration.
Although the platform includes an FPU for hardware-based floating point operations, we still apply int8 quantization to reduce memory usage with minimal accuracy loss, as can be seen in section~\ref{subsec:object_detection_performance}.
For quantization, we estimate the range of \textit{float32} values to map into \textit{int8}, using a subset of the training data to compute expected activation values for the neural network's layers.

\begin{figure}[tb]
\centering
\includegraphics[width=1.0\columnwidth]{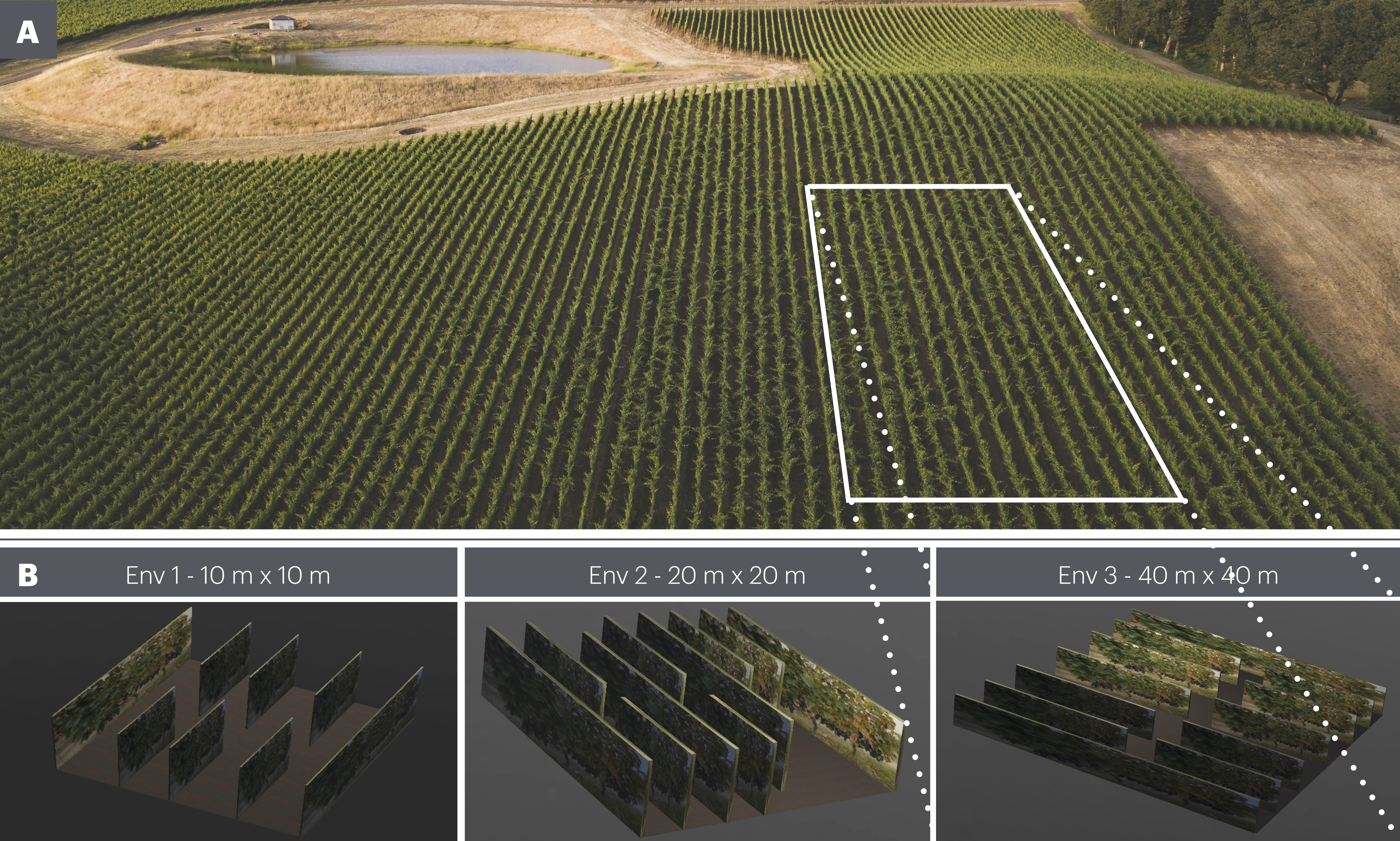}
\caption{A) Split of the vineyard. Each UAV explores one portion of the environment. B) Three different testing areas.}
\label{fig:webots_worlds}
\end{figure}

\begin{figure}[tb]
\centering
\includegraphics[width=1.0\columnwidth]{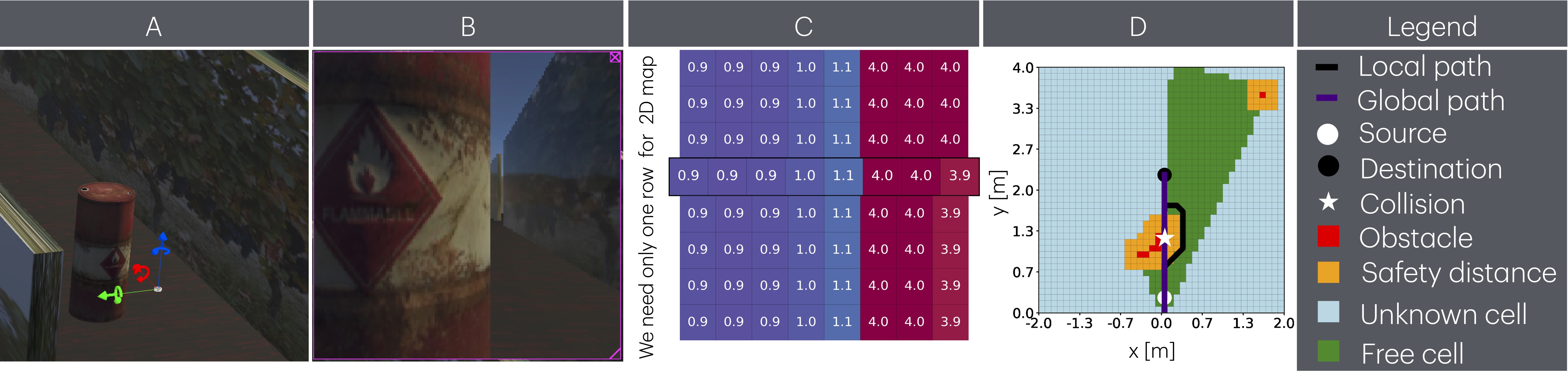}
\caption{A sample from the Webots simulator where the nano-UAV is facing an obstacle, with an external view in (A), onboard camera in (B), and ToF readings (in \SI{}{\meter}) in (C). D) 4$\times$\SI{4}{\meter} quantized (\SI{10}{\centi\meter} cells) local occupancy map obtained from the depth sensor. The map reports the local and the global path, with the collision point of the global path on the obstacle detected by the depth sensor.}
\label{fig:depth_to_occupancy}
\end{figure}

\subsection{Routing}
\label{subsec:routing}
\begin{wraptable}{R}{0.5\textwidth}
    \small\centering
    \caption{Cost of the path depending on the destination node occupancy and corresponding color in the occupancy map.}
    \addtolength{\tabcolsep}{10pt}
    \label{tab:occupancy_scheme}
    \resizebox{0.48\columnwidth}{!}{%
    \begin{tabular}{lll}
    \toprule
    \textbf{Destination node}&\textbf{Color scheme}&\textbf{Cost (c)}\\
    \midrule
    \textbf{Global path}&Violet&0\\
    \textbf{Free cell}&Green&25\\
    \textbf{Unknown cell}&Blue&50\\
    \textbf{Safety area}&Orange&75\\
    \textbf{Obstacle}&Red&$\infty$\\
    \bottomrule
    \end{tabular}
    }
\end{wraptable}

The routing algorithm that we use for obstacle avoidance relies on A*. 
This algorithm takes as input a 2D occupancy map, the source ($u_s$, $v_s$), and the destination ($u_d$, $v_d$), which are defined on the map with the horizontal and vertical coordinate ($u$, $v$).
The algorithm aims to find the optimal path (based on a cost metric) that connects the source and the destination, considering only feasible cells, i.e., cells that do not contain obstacles.
In this work, we employ two cost metrics to evaluate connections between nodes. 
The first metric considers only the Euclidean distance between nodes. 
The second combines the Euclidean distance with an additional cost $c$, which varies depending on the state of the next node. 
Nodes can belong to one of several categories: obstacle cells, which are identified as containing obstacles; safety area, which is within two cells of an obstacle; unknown cells, which are outside the line of sight of the ToF distance sensor; free cells, which are within the line of sight but not classified as obstacles or safety areas; and global path cells, which are part of the preplanned global path, provided they are not obstacles or within safety areas. 
These classifications, along with their associated costs, are detailed in Table~\ref{tab:occupancy_scheme}.


The cost is designed to guarantee safe obstacle avoidance, with global path cells having the lowest cost and obstacle cells having the highest. 
In a local map of size  $N\times M$  (e.g.,  N = 40  and  M = 40  in our setup), the cost of obstacle cells must exceed the aggregate cost of all non-obstacle cells to ensure they are never chosen by the path planning algorithm as such they are assigned with a cost value of infinity ($\infty$).
Safety area cells, located around obstacles, are assigned the next highest cost and are intended for traversal only when no alternative routes exist.  
Unknown cells, which represent areas outside the depth sensor’s field of view, are assigned a higher cost than free cells because they could potentially contain hidden obstacles or be near them (safety area cells). 
Free cells, apart from global path cells, are prioritized as the most suitable for traversal. 
However, global path cells take precedence over all others since they form part of the predefined route, which includes essential waypoints for inspection.
While the absolute values of the costs (e.g., 0, 25, 50, 75, and  $\infty$ ) can be adjusted, it is crucial to maintain the same ranking between cells' cost to ensure effective pathfinding.

The time that A* requires to find a solution that connects the source to the destination depends on the cost metric.
In fact, the cost metric is used to prune the number of paths that need to be updated.
A convex cost metric results in a lower number of updates since, at each update, the cost represents a global minimum.
A measure of the efficiency of a cost metric for the A* algorithm is named b*, i.e., the effective branching factor.
Given N, the number of nodes generated during the expansion, and d, the depth of the solution, the effective branching factor can be determined by solving the equation: $N=b^*+(b^*)^2+...+b^d$.

The multi-UAV path-planning problem is reduced to multiple single-UAV routing problems by splitting the entire field into several areas of maximum size 40$\times$\SI{40}{\meter} as depicted in Figure~\ref{fig:webots_worlds}-A; this is the maximum size that a single UAV can explore, at an average flight speed of \SI{1}{\meter/\second}, within its battery lifetime, i.e., $\sim$\SI{6}{\minute}.
To improve the variety of our testing areas, we create different sizes, as depicted in Figure~\ref{fig:webots_worlds}-B, i.e., Env 1 of size 10$\times$\SI{10}{\meter}, Env2 of size 20$\times$\SI{20}{\meter}, and Env 3 of size 40$\times$\SI{40}{\meter}.
The routing problem is split into two steps: a global and a local part.

The global routing problem is solved once for each UAV before the start of the exploration mission, and a predefined path through waypoints, which define the pose of the drone in 4 degrees of freedom (x, y, z, $\phi$), is created to explore all the rows of the vineyard. 
The global routing is performed on a quantized global occupancy map representing the entire environment of size $L_{global - map} \times L_{global - map}$ with $L_{global - map}=$\SI{10}{\meter},~\SI{20}{\meter} or~\SI{40}{\meter} depending on the environment as reported in Figure~\ref{fig:webots_worlds}-B.
The cells of the map are of side $L_{cell}=$10$\times$\SI{10}{\centi\meter}, which is equal to the side of the drone.
However, it considers only static objects that are present a priori in the map.
If the environment is static, i.e., no new obstacles appear on the map, the UAV can follow the predefined path and thus perform the exploration, but since the vineyard is a dynamic environment, new obstacles may appear between waypoints. 

To address this problem, we employ A* running entirely on the STM32 MCU available onboard the UAV. 
Our system is designed for 2D obstacle avoidance, limiting UAV movements to a horizontal plane, as vineyards primarily consist of open, horizontal areas. 
Although UAVs can avoid obstacles by moving vertically we want to apply the same obstacle avoidance algorithm to ground robots, that must eventually reach pest hotspots, which can only navigate in 2D. 
The need for a ground robot creates a critical requirement for largely unobstructed spaces to enable effective operation.
From the 8$\times$8 depth maps obtained from the ToF depth sensor, we extract the 4th row starting from the top, thus obtaining a 1$\times$8 array as shown in Figure~\ref{fig:depth_to_occupancy}-C. 
The eight distances of the 4th row, measured with the ToF sensor, are then back-projected on a quantized 2D local occupancy map, representing the top view of the environment in the FoV of the sensor per each instant.
The local occupancy map is a square of dimension $L_{map} \times L_{map}$ quantized with cells of side $L_{cell}$.
We define $L_{map}=$\SI{4}{\meter} since the maximum distance of the depth sensor is \SI{4}{\meter}.
Figure~\ref{fig:depth_to_occupancy}-D reports the occupancy map with the detected obstacles depicted in red as reported by the color scheme in Table~\ref{tab:occupancy_scheme}. 
Our system maps obstacles conservatively by marking an entire cell as occupied if any part contains an obstacle. 
This approach ensures the UAV avoids potential collisions with inaccurately mapped obstacles, as the method reliably designates all objects requiring avoidance as obstacle cells.

Our approach operates independently of time and is effective for static and dynamic obstacles by generating a new map for each depth map measurement. 
For dynamic obstacles as for static ones, we initiate local planning whenever an obstacle intersects the UAV’s current path. 
If an obstacle that previously caused replanning for an earlier depth map measurement overlaps the path again, the system triggers local planning, as it does not store a history of obstacle positions.
The pose of the UAV in the world frame for our work is assumed to be provided through tag-based localization or ultra-wideband systems~\cite{10257228}. 
Examples of maps are reported in Figure~\ref{fig:webots_worlds}-B.

\subsubsection{Simulation and onboard implementation}

\label{subsec:sim_env}
Webots\footnote{https://cyberbotics.com/\#webots} is an open-source, cross-platform simulator that integrates a model of the Crazyflie 2.1 nano-UAV, accurately simulating its dynamics and all of its sensors, including the camera. 
We use the simulator to develop and test our routing algorithms in the three different areas described in Section~\ref{subsec:routing}.
Figure~\ref{fig:depth_to_occupancy} shows a screenshot from the simulator, with the 3rd person view, the onboard camera view, the depth map obtained from the ToF, and the local occupancy map.

We implement the system in two versions: a Python-based version for performing simulations in Webots and a C-based version for deployment on the Crazyflie nano-UAV.
Since the local occupancy map is defined in the body frame of the UAV to perform a local routing, we consider the source always as the central cell on the $x$ axis, i.e., the cell ($\floor*{\frac{L_{map}}{L_{cell}}}$, 0) while as the destination we consider the next waypoint in the world frame projected on the local 2D occupancy map.
Figure~\ref{fig:depth_to_occupancy}-D depicts the source, the destination, the predefined path (red line), and the path that the UAV is currently following (black line), i.e., the locally planned path computed to avoid the obstacle.
The path obtained in the body frame of the UAV is then back-projected in the 3D space in the world frame to perform the tracking with the onboard controller. 
The back projection is performed considering a fixed flying altitude of \SI{1}{\meter}.

Since our UAV requires 4 degree-of-freedom waypoints (x, y, z, and $\phi$) to define a path, we consider the path in the world frame, adding the yaw for each step to orient the UAV with the ToF sensor continuously facing the direction of movement.
To obtain this behavior, we compute the yaw for each step as the $\phi=\arctantwo{(y_{d} - y_{s}, x_{d} - x_{s})}$ where s is the UAV current point and d is a point that is five steps forward on the path or the last point if the remaining path has less than five steps.

\section{Results}
\label{sec:results}
In this section, we report our system's results for both pest detection and the global+local planning task.
Furthermore, we provide an analysis of the benefits of employing a flying-ground cooperative system w.r.t. traditional ground pest treatment robots.

\subsection{Insect detector}

Following, we discuss the detection performance of our CNN for \textit{Popillia japoninca}, and we evaluate their performance on the selected embedded architecture.

\subsubsection{Object detection performance}
\label{subsec:object_detection_performance}

In this section, we assess the detection accuracy of each insect class, with a particular focus on identifying Popillia japonica. 
To evaluate the network's performance, we use the mAP, a standard object detection metric ranging from 0 to 1. 
A detection is considered correct if the Intersection over Union (IoU) between the predicted bounding box and the ground truth is at least 0.5, following the COCO standard~\cite{cocodataset}. 
The network is tested on a set of 660 samples with the most numerically accurate data type, i.e., \textit{float32}, and exploring variations in deployment \textit{float16} and \textit{int8} (quantized) formats.

\begin{wraptable}{R}{0.5\textwidth}
    \small\centering
    \caption{SSDLite-MobileNetV3 AP per each class of insect and per average bounding boxes dimension.}
    \label{tab:perclass_perbbox_mAP}
    \resizebox{0.48\textwidth}{!}{%
    \begin{tabular}{llll}
    \toprule
    \textbf{Configuration}& small & medium & large\\
    \midrule
    \textbf{Popillia japonica}&0.78&0.83&0.82\\
    \textbf{Cetonia aurata}&0.76&0.77&0.81\\
    \textbf{Phyllopertha horticola}&0.76&0.81&0.83\\
    \bottomrule
    \end{tabular}
    }
\end{wraptable}
When considering models in \textit{float32}, the SSDLite-MobileNetV3 (320$\times$240 input) reaches an mAP of 0.80. 
Additionally, when deploying the SSDLite-MobileNetV3 in \texttt{float16} to leverage the single-precision FPUs on the GAP9 SoC, we observe no significant mAP drop compared to its double-precision counterpart, i.e., mAP is equal to 0.80 in \texttt{float32} and \texttt{float16}. 
Lastly, when switching from \texttt{float32} to \texttt{int8} quantized version, the mAP reduction is less than 2\%.

We consider this performance satisfactory for the detection of \textit{Popillia japonica} as the insects tend to aggregate in clusters on host plants~\citep{eppo-popillia}, hence misdetecting roughly $1$ in $5$ specimens has a small impact on hotspots detection.
Furthermore, the detections performed with nano-UAVs are used to trigger the intervention in the hotspot's position of a larger ground robot that is not constrained to microcontroller-unit class devices but can host heavy power-demanding GPUs onboard. 
As such, it can perform detections using computationally intensive, yet more accurate, models such as the FasterRCNN-MobileNetV3-FPN network reported in Table~\ref{tab:other_models_params}.

Furthermore, we report in Table~\ref{tab:perclass_perbbox_mAP} the performance of our SSDLite-MobileNetV3 network per each class of insect and size of bounding boxes w.r.t. the entire image, i.e, small (10\%), medium (between 10\% and 25\%), and large (more than 25\%).
The performance of the network are consistent across all the configuration reported in Table~\ref{tab:perclass_perbbox_mAP} with the minimum average precision (AP) obtained by the Phyllopertha horticola in the case of small insects while the highest AP is achieved by the Phyllopertha horticola in the case of large bounding boxes; the model provides slightly more accurate predictions in the case of larger bounding box, i.e., medium and large, w.r.t. small ones.

\subsubsection{Embedded system deployment performance}

\begin{wraptable}{R}{0.5\textwidth}
    \small\centering
    \caption{GAP9 SoC configurations.}
    \label{tab:configurationsGAP9}
    \resizebox{0.48\textwidth}{!}{%
    \begin{tabular}{llll}
    \toprule
    \textbf{Configuration}&\textbf{Voltage [\SI{}{\volt}]}&\textbf{\textbf{Freq CL [\SI{}{\mega\hertz}]}}&\textbf{\textbf{Freq FC [\SI{}{\mega\hertz}]}}\\
    \midrule
    \textbf{Min power}&0.65&150&150\\
    \textbf{Max efficiency}&0.65&240&240\\
    \textbf{Min latency}&0.80&370&370\\
    \bottomrule
    \end{tabular}
    }
\end{wraptable}

In this section, we conduct a comprehensive evaluation of the performance of all models deployed on the GAP9Shield. 
Table~\ref{tab:configurationsGAP9} details three GAP9 SoC configurations, with voltages ranging from~\SI{0.65}{\volt} to~\SI{0.8}{\volt}, enabling clock speeds of up to~\SI{370}{\mega\hertz}. 
For our experiments, we adopt the \textit{maximum efficiency} configuration, which provides optimal battery life while maintaining the target throughput.

Table~\ref{tab:results} presents a detailed analysis of each model’s memory requirements, inference latency per image, and total power consumption, including both compute unit and off-chip memory usage. 
The model is evaluated using different data types (\texttt{float16} and \texttt{int8}).
The peak memory usage is determined based on the network layer with the highest memory demand for input tensors, weights, and outputs. 
NN-Tool on the GAP9 can efficiently manage memory across the~\SI{1.6}{\mega\byte} on-chip L2 and up to~\SI{32}{\mega\byte} of off-chip memory.

We deploy the SSDLite-MobileNetV3 with a fixed input size of 320$\times$240$\times$3 (RGB) across three configurations: \textit{i}) \texttt{float16} running on general-purpose CL with FPU hardware support, \textit{ii}) \texttt{int8} quantized running on the CL, and \textit{iii}) \texttt{int8} using the NE16 convolutional accelerator. 
The \texttt{float16} version, while supported by the hardware FPU, requires the most memory to store all the weights ($\sim$\SI{7}{\mega\byte}) and the most peak memory during execution (at most $\sim$\SI{3.6}{\mega\byte}). Furthermore, it is the slowest, processing at~\SI{2.1}{frame/\second} with~\SI{40.6}{\milli\watt} of power consumption. 
By contrast, the \texttt{int8} version running on the NE16 accelerator requires only~\SI{3.4}{\mega\byte} of memory for all the weights while requires a peak ~\SI{1.8}{\mega\byte} memory for the execution, reaching~\SI{6.8}{frame/\second} while consuming~\SI{34}{\milli\watt}.
Table~\ref{tab:results} reports the peak memory required during the execution of the networks.
We do not deploy the \texttt{float32} version of the network since the GAP9 SoC does not have hardware support for this type of computations and, as such, this will result in a sub-optimal performance, in terms of throughput and power consumption, due to the soft-float emulation required which is not justified due to the iso-accuracy w.r.t. the \texttt{float16} version of the network.

\begin{table*}[t]
    \small
    \caption{Latency, power consumption, and mAP of all networks on the GAP9 SoC. 
    }
    \label{tab:results}
    \resizebox{\linewidth}{!}{%
    \begin{tabular}{llcccccc}
    \toprule
    \textbf{Network} & \textbf{Data type} & \textbf{Input size} & \textbf{MCU} & Memory [\SI{}{\kilo\byte}] * & \textbf{Latency [\SI{}{\milli\second}]} & \textbf{Total power [\SI{}{\milli\watt}]} & \textbf{mAP} \\
    \toprule
    SSDLite-MobileNetV3 & \texttt{float16} & 320$\times$240$\times$3 & GAP9 (CL) & 3622 & 462 & 41 & 0.80\\
    SSDLite-MobileNetV3 & \texttt{int8} & 320$\times$240$\times$3 & GAP9 (CL) & 1811 & 249 & 31 & 0.79\\
    SSDLite-MobileNetV3 & \texttt{int8} & 320$\times$240$\times$3 & GAP9 (NE16) & 1811 & 147 & 34 & 0.79\\
    \bottomrule
    \vspace{0.01pt}
    \end{tabular}
    }
    \footnotesize{* The memory footprint includes the input and output tensors and weights of the largest network layer and the input image.}
\end{table*}

\begin{figure}{}
\centering
\includegraphics[width=0.5\textwidth]{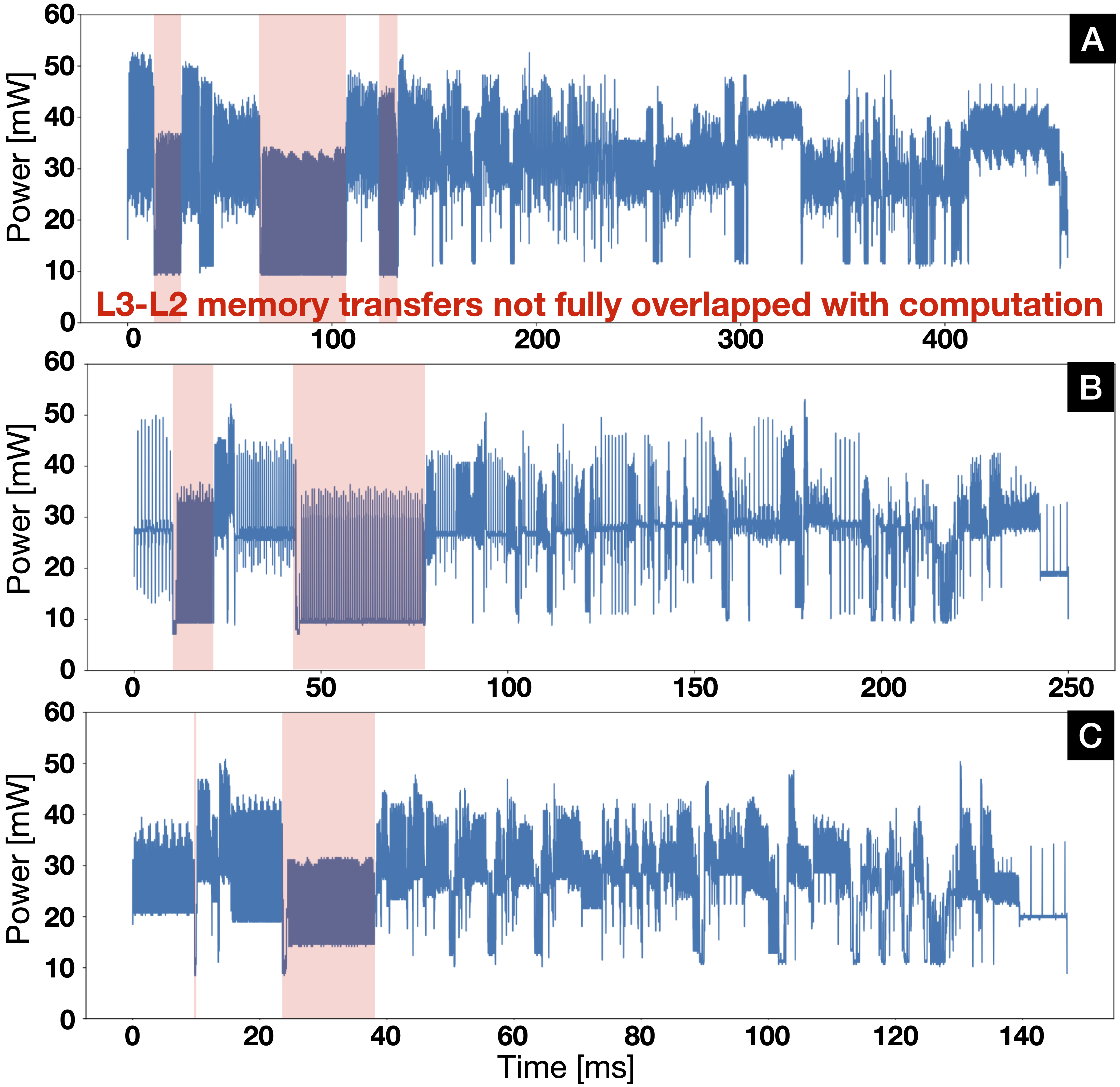}
\caption{Power waveforms for the three deployed networks on the GAP9, i.e., A) float16, B) int8 running on the CL, and C) int8 running on the NE16 accelerator.}
\label{fig:powerGAP9}
\end{figure}

Figure~\ref{fig:powerGAP9} displays the power consumption waveforms for the GAP9 SoC under three conditions: \textit{i}) \texttt{float16}, \textit{ii}) \texttt{int8} on the general-purpose multicore cluster, and \textit{iii}) \texttt{int8} on the NE16 accelerator, shown in panels A, B, and C, respectively. 
All three configurations exhibit short non-overlapped memory transfers (up to~\SI{41}{\milli\second} in total), particularly at the start of execution when tensor movements between memory layers do not overlap with computation. 
These non-overlapped memory transfers decrease as tensor sizes shrink during network execution.

\subsection{Routing}

We evaluate our routing w.r.t. its effectiveness in visiting the predefined waypoints and from a latency standpoint.

\subsubsection{Waypoints visited}

\begin{table}[tb]
    \small\centering
    \caption{Obstacle avoidance performance of four different algorithms, i.e., blind (B), random (R), local routing weighted (W), and local routing shortest (S), in three different environments with changing number of obstacles. Performance measured in percentage of waypoints reached. R considers a position within the waypoint if the distance is less than~\SI{30}{\centi\meter}. }
    \label{tab:changing_obstacles}
    \resizebox{0.7\columnwidth}{!}{%
    \begin{tabular}{lcccccccccccc}
    \toprule
    \multirow{2}[3]{*}{\shortstack[c]{\textbf{Area}\\\textbf{occupied [\%]}}} & \multicolumn{4}{c}{\textbf{Env 1 - 10$\times$\SI{10}{\meter}}}&  \multicolumn{4}{c}{\textbf{Env 2 - 20$\times$\SI{20}{\meter}}} &  \multicolumn{4}{c}{\textbf{Env 3 - 40$\times$\SI{40}{\meter}}}\\
    \cmidrule(lr){2-5} \cmidrule(lr){6-9} \cmidrule(lr){10-13}
    & B & R & W & S & B & R & W & S & B & R & W & S\\
    \midrule\textbf{0}&100&84&\textbf{100}&100&100&44&\textbf{100}&100&100&13&\textbf{100}&100\\
\textbf{1}&88&83&\textbf{100}&53&56&43&\textbf{100}&65&30&10&\textbf{100}&32\\
    \textbf{2}&47&81&\textbf{100}&53&35&42&\textbf{100}&65&21&9&\textbf{80}&32\\
    \textbf{5}&37&71&\textbf{100}&20&36&41&\textbf{100}&54&20&9&\textbf{77}&32\\
    \textbf{10}&20&61&\textbf{100}&20&12&29&\textbf{100}&65&5&8&\textbf{50}&10\\
    \bottomrule
    \end{tabular}
    }
\end{table}

We evaluate our system using the three different maps described previously in Section~\ref{subsec:sim_env}: Env 1, Env 2, and Env 3.
We perform one run for each map represented in Figure~\ref{fig:webots_worlds}, gradually increasing the number of obstacles.
The number of obstacles is expressed as a percentage of the total area of the map, and we test five different configurations: 0\%, 1\%, 2\%, 5\%, and 10\%. 
The obstacles are cylindric barrels of~\SI{1}{\meter\squared} randomly positioned in the map without overlapping waypoints and checking that there is always at least one path available to reach the next waypoint of the path.
In this experiment, we compare all the routing methods described before each evaluated on every randomly generated configuration of obstacles.
The results of the experiment are reported in Table~\ref{tab:changing_obstacles}.
The algorithms reported are:
\begin{itemize}
  \item Blind (B), i.e., the UAV follows the predefined path without considering the obstacle seen by the ToF sensor. 
  \item Random (R), the UAV performs a path composed of segments with uniformly sampled random orientation in [$-\pi, \pi$] and length between 0 and $min(dist,\SI{4}{\meter})$  where $dist$ represents the minimum distance retrieved at that instant by the ToF sensor.
  The random path is executed until the end of the battery, i.e.,~\SI{5}{\minute} and considering~$\sim$\SI{1}{\minute} to return to the starting point.
  \item Weighted (W) uses local routing with the weighted cost metric.
  \item Shortest (S) uses local routing, considering only the distance between visited cells as a cost metric.
\end{itemize}

As expected, the best-performing routing algorithms are the global ones in all environments; in fact, they are always able to find a solution for the considered path. 
The use of a global occupancy map allows the finding of globally optimal solutions for the paths and provides feasible path solutions even if local solutions are not available due to the presence of obstacles in the local occupancy map.
If we limit the knowledge to a local map, i.e., the map created from the ToF sensor data, the local routing with the W cost metrics achieves the best results in terms of the percentage of waypoints visited w.r.t. the total number of waypoints on the map.
It achieves up to 100\% of waypoints visited in Env1 and Env2 with all obstacle configurations, while in Env 3, it reaches the end of the path, i.e., 100\% of waypoints visited, only in the case of 0\% and 1\% of the total area occupied by obstacles.
Further increasing the percentage of occupied area leads to a decrease in the portion of path completed bottoming at 50\% in the case of 10\% of area occupied, i.e., 160 obstacles on the map.
Still, the routing method provides an improvement over the B and R baselines that reach up to 45\%.

The S cost metric underperforms in local routing compared to the weighted (W) cost metric. In particular, in our smallest environment, Env 1, the S cost metric yields the lowest performance among all methods. Paths generated with the S metric are more likely to encounter blockages when obstacles are nearby and partially obscured. Although the W cost metric can occasionally face similar issues, it does so less frequently under the same conditions (environment and obstacle positions). Local routing with the W metric demonstrates higher reliability, and blockages are encountered less often than with the S metric. This limitation can be effectively mitigated by introducing random yaw adjustments and forward movements toward a free direction when a blockage occurs.

\begin{figure}[tb!]
\centering
\includegraphics[width=1.0\columnwidth]{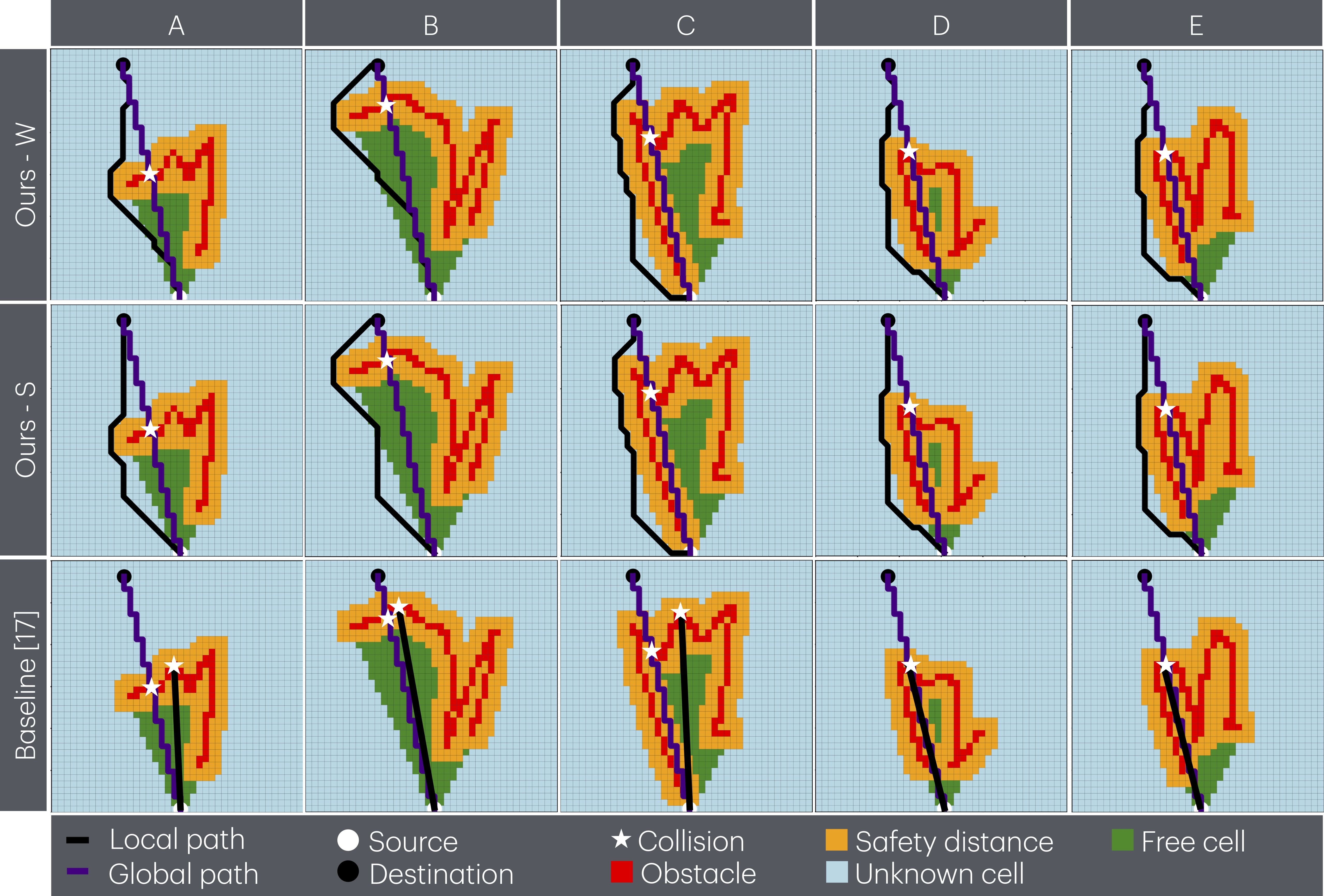}
\caption{Local routing: plannings with five different obstacles. The cost metric for each replanning is reported on the left. A* refers to the path replanned with the A* algorithm.}
\label{fig:paths_obstacle}
\end{figure}

Figure~\ref{fig:paths_obstacle} provides five examples of obstacle avoidance of irregularly shaped non-uniform obstacles reported within the local occupancy map computed onboard the UAV. 
These non-uniform obstacles may be present in a real-world vineyard; Figure~\ref {fig:paths_obstacle} highlights the capabilities of our algorithm to avoid nonconvex obstacles.
We provide results with our two cost metrics, i.e., the W and the S path cost metrics.
Figure~\ref{fig:paths_obstacle}-A and~\ref{fig:paths_obstacle}-B displays the difference between the two algorithms at the beginning of the path, near the source (white), where the local routing with the W cost metric provides a path that is equal to the pre-planned path and traverses the free area (green) which is desirable since the unknown cells may contain obstacles that are outside the FoV of the sensor.
Local routing with the S cost metric, instead, provides a solution that traverses the uncertain area with the risk of finding an obstacle outside the depth sensor's FoV.
Figure~\ref{fig:paths_obstacle}-C reports the same path for both cost metrics since traversing the free area does not lead to a feasible solution in the case of the W cost metric.
Figure~\ref{fig:paths_obstacle}-D and~\ref{fig:paths_obstacle}-E displays the same difference of~\ref{fig:paths_obstacle}-A and~\ref{fig:paths_obstacle}-B but at the end of the path, near the destination point (black).

The last row of Figure~\ref{fig:paths_obstacle} reports local planning performed with a baseline reactive local navigation algorithm~\cite{6630610}.
This algorithm chooses the heading and the speed of the robot depending on the obstacles in the environment.
The heading is chosen to minimize the distance from the endpoint of the current movement to the destination.
The baseline~\cite{6630610} struggles when the robot’s distance sensors encounter non-convex irregular obstacles that fully occupy their field of view (FoV). In such cases, the robot may become stuck in local minima and unable to avoid the obstacle. 
By contrast, our algorithm supports path planning beyond the FoV, even when obstacles entirely obstruct it.
To conclude, we provide a video\footnote{\href{https://github.com/idsia-robotics/A-star-path-planning-nano-drones}{https://github.com/idsia-robotics/A-star-path-planning-nano-drones}} with an experiment of obstacle avoidance done in simulation using our A*-based algorithm with the W cost.

\subsubsection{Onboard computing time}

\begin{wrapfigure}{R}{0.5\textwidth}
  \centering
    \includegraphics[width=0.48\textwidth]{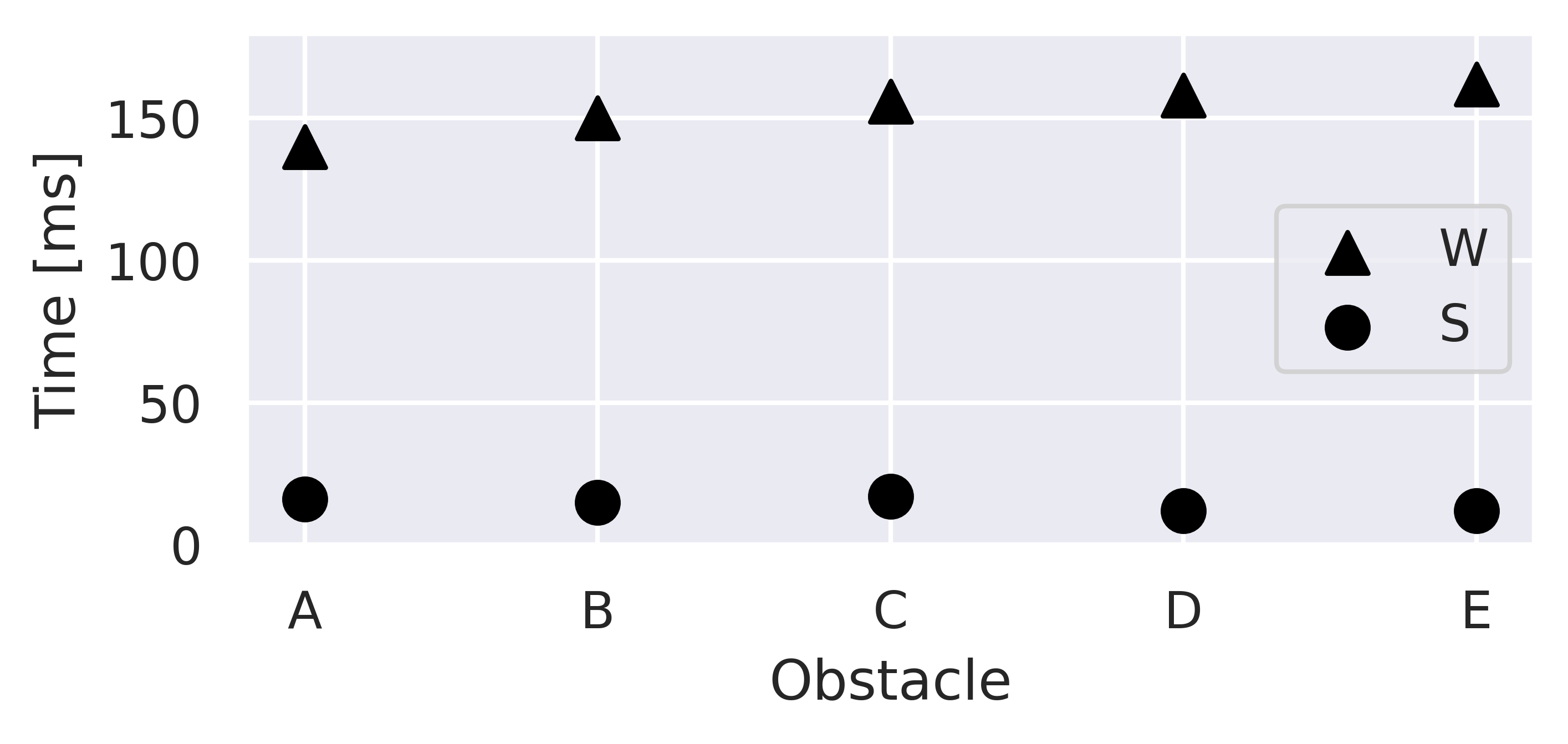}
  \caption{Time [\SI{}{\milli\second}] to compute A* path with the STM32 MCU available onboard the Crazyflie 2.1 with five different obstacles, depicted in Figure~\ref{fig:paths_obstacle}, using the local routing (4$\times$\SI{4}{\meter} maps) with two cost metrics weighted and shortest path.}
  \label{fig:time_a_star}
\end{wrapfigure}

Figure~\ref{fig:time_a_star} reports the onboard measurements of the time to compute a solution for the five different obstacles shown in Figure~\ref{fig:paths_obstacle}. 
In particular, the W cost metric requires up to $12\times$ the time required to find a solution with the S cost metric.
This is the case since the S cost metric allows the pruning of the paths that need to be saved because the cost depends only on the distance and thus has a branching factor b* that is lower than the W cost metric.
This results in an expansion phase with a lower number of nodes in the case of the S cost metric, which consequently leads to a proportionally lower amount of memory required w.r.t. the W cost metric. 

However, given the higher performance in the obstacle avoidance task achieved by the W cost metrics, reported in Table~\ref{tab:changing_obstacles}, and considering that the A* algorithm is used only in the case an obstacle is detected on the pre-planned path, thus achieving real-time control performance, the most promising approach to perform obstacle avoidance onboard the Crazyflie 2.1 is the local routing with W cost metric.
The onboard routing algorithm runs on the STM32 MCU available onboard, which has a peak power consumption of \SI{138}{\milli\watt}@\SI{168}{\mega\hertz}.

\subsection{Complete system performance}

In this section, we evaluate the performance of our flying-ground cooperative system in the treatment of pest insects in a 200$\times$\SI{200}{\meter} vineyard.
The vineyard is initially split into 25 areas of 40$\times$\SI{40}{\meter}, and one UAV per area executes an exploration looking for the Popillia japonica that takes only~$\sim$\SI{5}{\minute} considering an average speed of the UAV of~\SI{1}{\meter/\second}.
After the exploration phase is completed, a ground robot capable of performing pest control visits all the points where an insect has been detected.

\begin{wrapfigure}{R}{0.5\textwidth}
\centering
\includegraphics[width=0.48\textwidth]{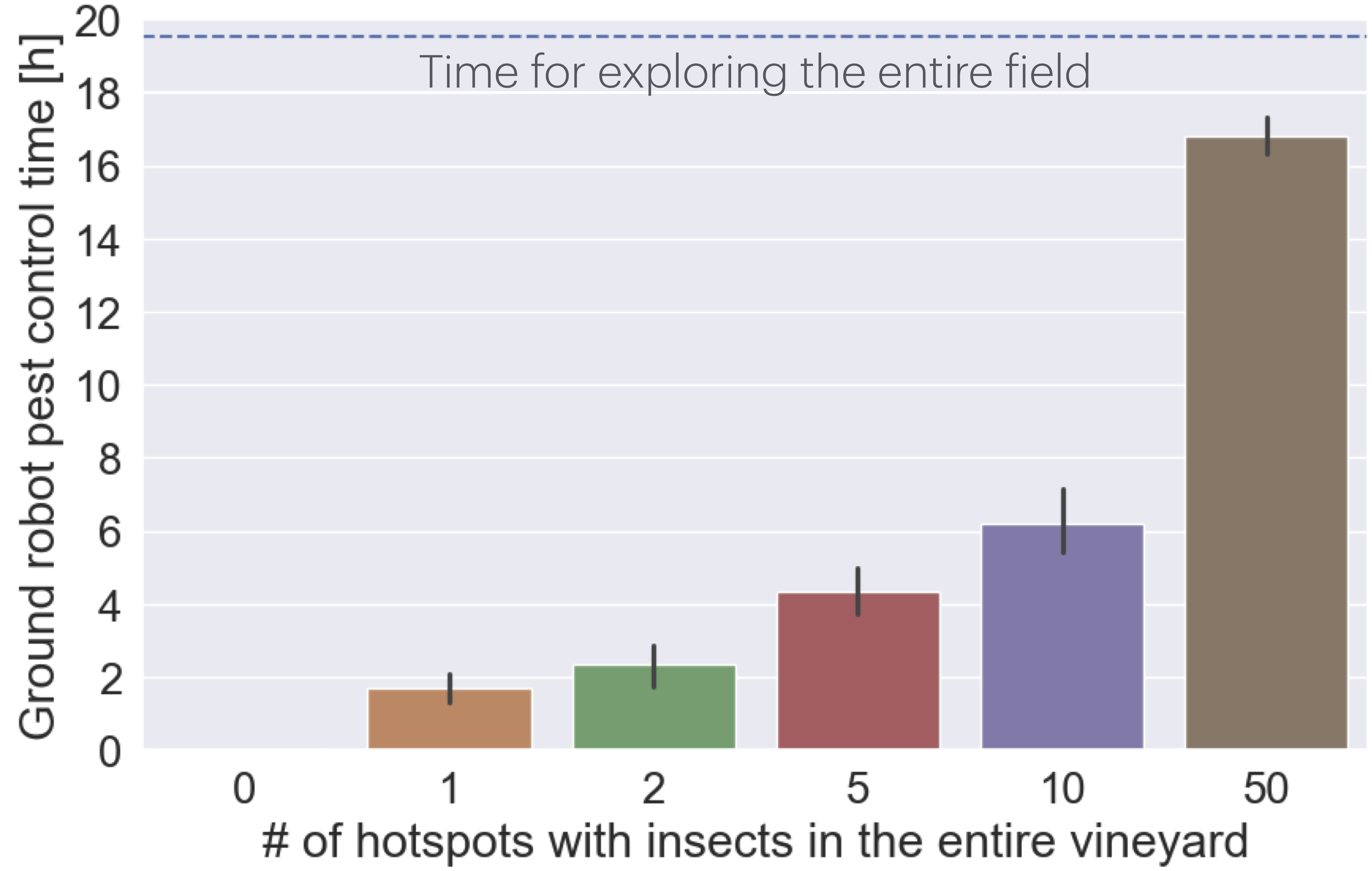}
\caption{The time, in hours, a ground robot (speed \SI{0.2}{\meter/\second}~\cite{9177181}) needs to reach all the detected insects in a vineyard. The experiments are done knowing the exact location of $N$ hotspots of insects, thanks to the exploration done with UAVs and without prior information, i.e., no prior exploration line.}
\label{fig:insects_ground_robot}
\end{wrapfigure}

To evaluate the improvement in time achieved by the ground robot to perform pest control over the vineyard, we analyze two conditions: 
\begin{itemize}
    \item no prior exploration with nano-UAVs, i.e., the ground robot needs to explore the entire field;
    \item prior exploration with nano-UAVs with a variable number ($N \in [0, 50]$) of randomly positioned hotspots in the vineyard.
\end{itemize}
The prior exploration with the nano-UAVs creates a map of the hotspots in the field that the ground robot explores in a second step to perform pest control.
The improvement is evaluated by comparing the time required by the ground robot to perform pest control on the entire vineyard, i.e., no prior exploration, w.r.t. the case the ground robot performs the pest control only in the hotspots found by the UAVs.
We report the average time required by the ground robot to perform pest control over five runs per configuration in Figure~\ref{fig:insects_ground_robot}.

The shortest path that connects the affected points is computed using the global routing algorithm with A*, simplifying the complexity of the numerical problem thanks to the structure of the vineyard.
Without the exploration done by the UAVs, the total pest control time results in~$\sim$\SI{20}{\hour} independently from the number of hotspots in the vineyard.
With the exploration instead, the time depends on the number of hotspots and ranges between 0 and~\SI{17}{\hour} in the case of 50 clusters of insects in the vineyard.
Our system gives the highest benefit when the number of hotspots is low, which is the normal operating condition in a healthy vineyard since an increasing number of insect hotspots results in the ground robot having to travel the entire vineyard.
In the case of zero hotspots detected, the ground robot is not operated, resulting in the maximum working time saving of $\sim$\SI{20}{\hour}.

The applicability of the proposed solution is determined by the convergence of technical limitations, environmental conditions, and characteristics of the target insect(s). Different insect species fly in different conditions and at different times of the day. For \textit{Popillia japonica}, the greatest flight activity is reported to be on clear days and when the temperature is between 29°C and 35°C, relative humidity greater than 60\% and wind is lower than 20 km/h \cite{eppo-popillia}. These conditions perfectly fit with the technical limitations of the drones for what concerns constraints on flight conditions (e.g., max wind speed) and lightning for the camera.

\section{Discussion and system limitations}

Our system employs compact nano-UAVs for both greenhouses and outdoor applications. 
Indoor environments provide ideal conditions with no weather constraints, while outdoor functionality is limited by weather conditions (e.g., wind speeds up to ~\SI{3.5}{\meter/\second}~\cite{9811834} and no rain). 
Though primarily demonstrated for pest detection, the system can also be applied to tasks like dry plant detection, crop monitoring, and counting. 
For Popillia japonica, optimal deployment conditions are sunny days with light winds and temperatures around \SI{29}{\celsius}~\cite{eppo-popillia} that fit the nano-UAVs' ideal operating conditions.

\begin{wrapfigure}{R}{0.48\textwidth}
\centering
\includegraphics[width=0.48\textwidth]{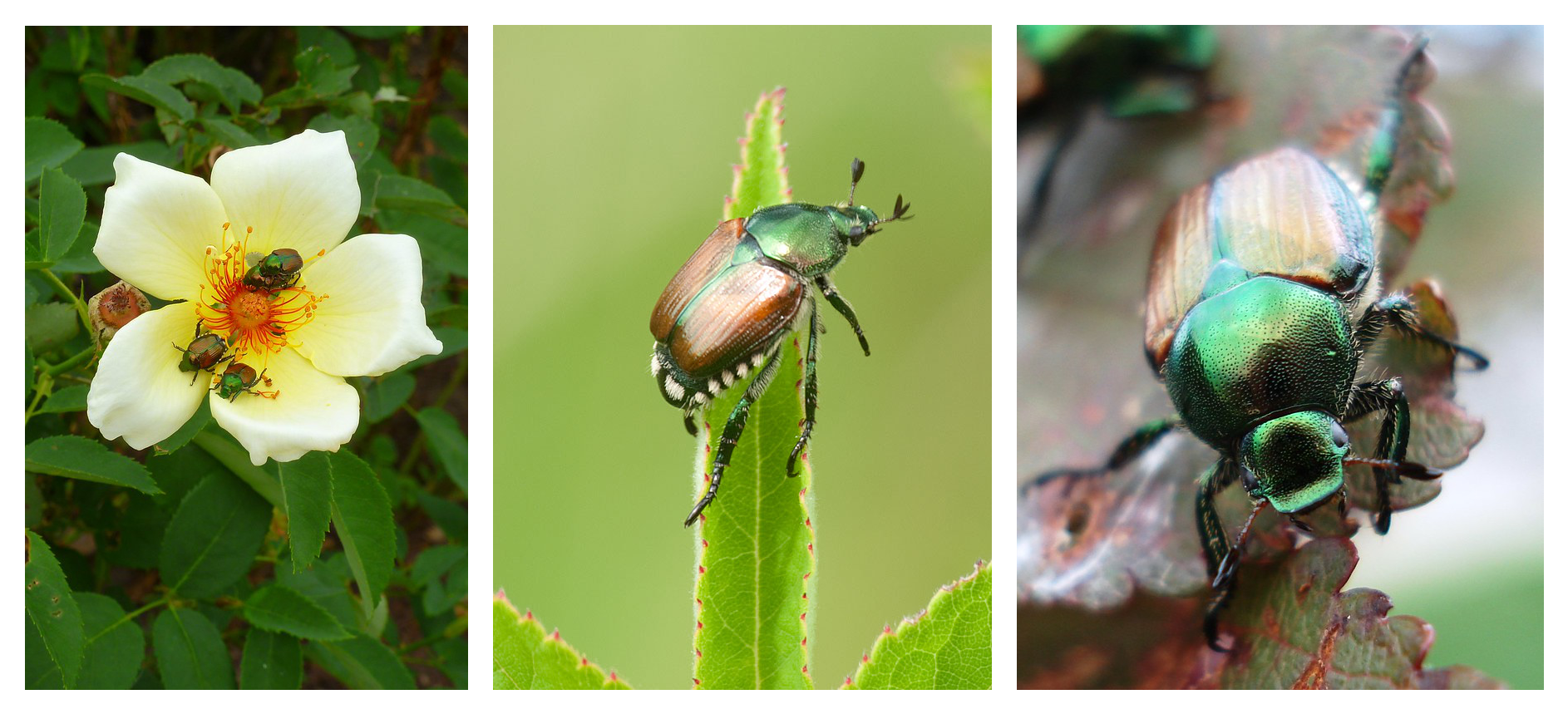}
\caption{Example images depicting \textit{Popillia japonica} specimens at different scales, relative to image size. From left to right, bounding boxes occupy, on average, 0.9\%, 23.9\% and 41.9\% of the image's total pixels.}
\label{fig:popillia_sizes}
\end{wrapfigure}

Nano-UAVs are of particular interest for pest detection because they have a lower environmental impact than standard-sized drones that are currently used for this application.
In fact, nano-UAVs produce noise up to 40 dB~\cite{10.1145/3666025.3699337} while their bigger counterpart reaches up to 75 dB~\cite{ijerph18126202}, as such nano-UAVs provide an interesting solution that reduces the impact on the environment.

We test the system across varying hotspot densities (0 to 50) and three crop arrangements (environments 1, 2, and 3), highlighting its advantages over ground robots for early pest detection. 
Performance is influenced more by obstacle density than by the overall environment layout, thanks to the reliance on local path planning for obstacle avoidance.
We now analyze in detail the limitations of our insect detector and of the routing algorithm.

\subsection{Insect detector}

The dataset used in our work contains images gathered online rather than images taken directly from a nano-drone. 
This implies the presence of both pictures where the insects appear in close proximity, and pictures where the insects are much smaller relative to image size. 
On average, target insects cover 17.1\% of an image’s total pixels (8.9\% for \textit{Popillia japonica}, 19.8\% for \textit{Phyllopertha horticola}, 22.5\% for \textit{Cetonia aurata}). 
Figure~\ref{fig:popillia_sizes} provides a visual reference for this. From left to right, bounding boxes occupy, on average, 0.9\%, 23.9\% and 41.9\% of the image's total pixels.
We believe nano-drones can produce similar images, given their small size and capability of flying between vineyards' rows, close to vines. 
However, we point out that a real-world deployment would likely benefit from a model trained directly on images acquired by the drones during exploration.

\subsection{Routing}

\begin{figure}[tb!]
\centering
\includegraphics[width=1.0\columnwidth]{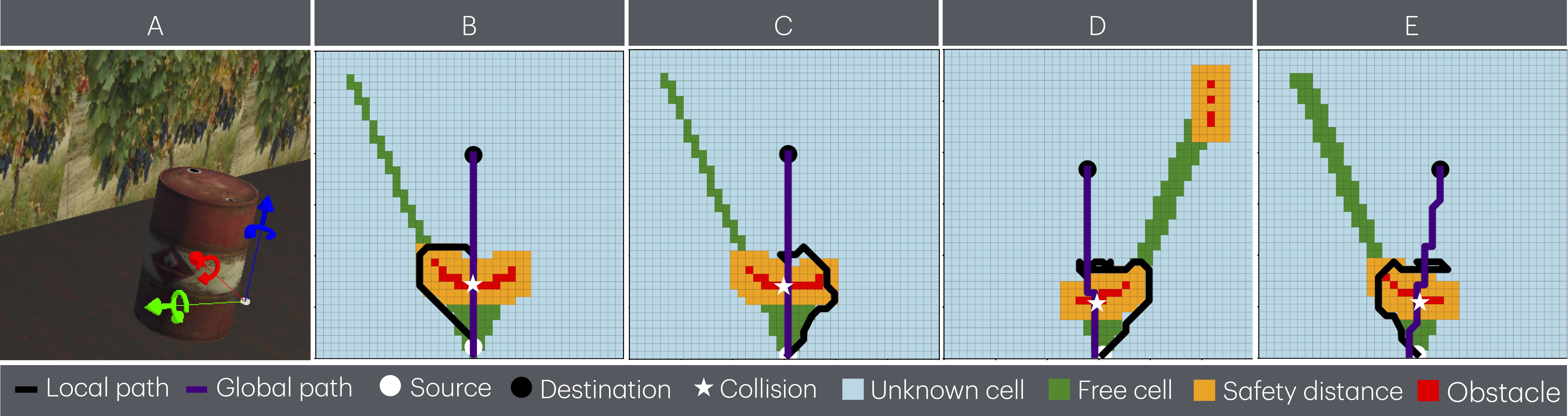}
\caption{An obstacle covers the entire depth sensor FoV, causing the blockage.}
\label{fig:deadlock}
\end{figure}

Blockages are a typical problem when relying on local planning strategies, they can occur when obstacles cover the entire FoV of the depth sensor, as reported in Figure~\ref{fig:deadlock}.
In fact, in this condition, the local planner provides a solution that passes through the uncertain region of the map.
However, when the UAV starts moving toward the uncertainty region, the depth sensor detects the presence of an obstacle that intersects the new locally planned path.
This causes a new iteration of the local planner, which provides a new solution that belongs to the uncertainty region that will result in a collision as soon as the UAV moves towards it.

The first solution involves maintaining a record of the surroundings based on the current depth measurement. 
This approach uses the same local planning algorithm proposed in this work, namely A*, applied to a local map that includes the current depth measurement along with all previous measurements within a 4$\times$\SI{4}{\meter} area used for local planning. 
This method gradually maps objects larger than the sensor’s FoV, which might otherwise obscure entirely the area ahead of the sensor and cause blockages. 
While this approach reduces the frequency of blockages, it does not eliminate them, as objects exceeding \SI{4}{\meter} in size can still lead to blockages in the current implementation.
Other local solutions to mitigate the blockages issue rely on reinforcement learning and swarm cooperation. 
Still, in our use case, that does not involve communication between UAVs and limits the knowledge of the map to a local instance of 4$\times$\SI{4}{\meter} obstacles that occlude the entire FoV may always result in blockages.
To avoid the blockage issue, a reliable solution is to perform obstacle avoidance on the global map, which cannot be done on our nano-UAVs due to the platform's computational constraints.

\section{Conclusions}

This work presents the building blocks for a novel, efficient transportation system applied to a pest detection and control scenario in vineyards, relying on flying and ground robots. 
We propose an implementation of the pest detection system that runs onboard the Crazyflie 2.1 nano-UAV on the ultra-low power multi-core GWT GAP9 SoC.
Due to the limited resources available on this platform, we explore and deploy a CNN-based detection system, i.e., the SSDLite-MobileNetV3 CNN (\SI{584}{\mega MAC/inference}), scoring an mAP of 0.79 with a throughput of~\SI{6.8}{frame/\second} at~\SI{33}{\milli\watt} on the GAP9 SoC.

We integrate the CNN-based insect detector with an obstacle avoidance algorithm running onboard the Crazyflie 2.1 nano-UAV to allow the autonomous exploration of vineyards.
Our local routing A*-based obstacle avoidance algorithm is able to reach up to 100\% of the planned waypoints, avoiding all the obstacles in two out of three environments with increasing complexity (from 0 to 10\% of the entire area covered with~\SI{1}{\meter\squared} obstacles).

If compared to the pre-planned path (B) and to the random (R) explorer baselines, our local routing algorithm (W) increases the number of waypoints visited within the battery life of the UAV between 16\% in the smallest environment, i.e., a~10$\times$\SI{10}{\meter} vineyard, to 90\% in our~40$\times$~\SI{40}{\meter} environment. 
The algorithm achieves real-time performance with planning requiring less than \SI{170}{\milli\second} on the STM32 MCU available on the nano-UAV while performing real-time flight control tasks.
Our multi-UAV system, using a swarm of 25 UAVs to explore a~200$\times$\SI{200}{\meter} vineyard, allows us to save up to~$\sim$\SI{20}{\hour}, i.e., 100\% of the time if no insects are detected, to perform pest control, paving the way to fully autonomous precise and effective pest control with an efficient transportation system applied to vineyards.

\bibliographystyle{ACM-Reference-Format}
\bibliography{biblio}

\appendix

\end{document}